\documentclass[10pt,letterpaper,twocolumn]{article}
\usepackage{preprint}
\usepackage[T1]{fontenc}
\usepackage[utf8]{inputenc}

\usepackage{amsmath,amssymb,amsthm}
\usepackage{array,graphicx,booktabs,multirow,tabularx}
\usepackage{cite,url,textcomp}
\usepackage{xurl}
\usepackage{tikz}
\usetikzlibrary{arrows.meta,positioning,fit,backgrounds}
\usepackage{placeins}
\usepackage{microtype}
\usepackage{colortbl}
\usepackage[normalem]{ulem}
\usepackage[export]{adjustbox}
\usepackage[font=small,labelfont=bf]{caption}
\usepackage[hidelinks,hyperfootnotes=false]{hyperref}

\allowdisplaybreaks
\author{%
  Rong Huang\textsuperscript{\textdagger},
  Yi Xiong\textsuperscript{\textdagger},
  Yuan-Yuan Cheng\textsuperscript{*}, and
  Xiao-Ming Fu\textsuperscript{*}\\[4pt]
  \normalfont\small University of Science and Technology of China\\
  \normalfont\small Hefei, Anhui 230026, China\\[3pt]
  \normalfont\small\texttt{sa25001099@mail.ustc.edu.cn},
  \normalfont\small\texttt{1440017889@qq.com}\\
  \normalfont\small\texttt{chyy@mail.ustc.edu.cn},
  \normalfont\small\texttt{fuxm@ustc.edu.cn}%
}

\title{Hierarchical Low-Rank Reparameterization for Parameter-Efficient Fine-Tuning}
\date{}
\hypersetup{
  pdftitle={Hierarchical Low-Rank Reparameterization for Parameter-Efficient Fine-Tuning},
  pdfauthor={Rong Huang; Yi Xiong; Yuan-Yuan Cheng; Xiao-Ming Fu},
  pdfkeywords={Hierarchical reparameterization, large language models, low-rank adaptation, parameter compression, parameter-efficient fine-tuning}
}

\begin{document}
\twocolumn[
\begin{@twocolumnfalse}
\maketitle
\begin{center}
  {\small\textsuperscript{\textdagger}Rong Huang and Yi Xiong contributed equally to this work and share first authorship.\\
  \textsuperscript{*}Corresponding authors: Yuan-Yuan Cheng and Xiao-Ming Fu.}
\end{center}
\begin{abstract}
Parameter-efficient fine-tuning (PEFT) methods adapt large models by introducing a small number of trainable parameters. 
However, we find that these parameters still contain redundancy and can be further compressed while maintaining strong task performance.
To this end, we propose a novel hierarchical low-rank reparameterization method. Its core idea is to reorganize the trainable parameters of PEFT methods
into an approximately square matrix and then express it via a low-rank decomposition, which can be applied recursively. 
This forms a multi-level structure with factors compressed level by level, drastically reducing the number of trainable parameters. 
In our experiments, we combine this reparameterization with LoRA, DoRA, MoSA, and RaSA. 
We evaluate commonsense reasoning on LLaMA-3.1-8B and Qwen2.5-7B, with additional GSM8K math reasoning experiments on LLaMA-3.1-8B.
Under comparable trainable-parameter budgets,
our method combined with LoRA or DoRA achieves better performance. 
Furthermore, when the parameter count of the original methods is compressed by approximately an order of magnitude, our method can still maintain competitive task performance and even surpass the corresponding baselines in average accuracy across the eight commonsense reasoning benchmarks.

\end{abstract}
\vspace{0.25cm}
\end{@twocolumnfalse}
]

\begingroup
\renewcommand{\thefootnote}{}
\footnotetext{This work has been submitted to the IEEE for possible publication.
Copyright may be transferred without notice, after which this version may no longer be accessible.}
\endgroup

\section{Introduction}\label{sec:introduction}
Large-scale pretrained models have demonstrated remarkable capabilities across a wide range of tasks, but adapting these models to downstream applications remains challenging due to their enormous parameter scale~\cite{hu-etal-2023-llm}. 
Parameter-efficient fine-tuning (PEFT) addresses this challenge by freezing the pretrained backbone and introducing a small number of trainable parameters~\cite{pmlr-v97-houlsby19a,lester-etal-2021-power,li-liang-2021-prefix,hu2022lora}. 
Existing PEFT methods have substantially reduced the cost of model adaptation~\cite{gao2024parameterefficient,han2024parameterefficientfinetuninglargemodels}, but the introduced trainable parameters may still contain internal redundancy that has not been fully explored.

Low-Rank Adaptation (LoRA) has become one of the most widely adopted PEFT methods due to its simplicity and effectiveness~\cite{hu2022lora}. LoRA represents weight updates using a low-rank parameterization, in which only additional trainable parameters are optimized while the pretrained weights remain frozen. 
Many LoRA variants have been proposed to improve parameter efficiency, optimization, initialization, and adaptation capacity~\cite{zhang2023adaptive,hayou2024loraefficientlowrank,NEURIPS2024_db36f4d6,liu2024dora,jiang2024morahighrankupdatingparameterefficient}. 
For example, AdaLoRA~\cite{zhang2023adaptive} improves rank allocation, PiSSA~\cite{NEURIPS2024_db36f4d6} and EVA~\cite{NEURIPS2025_41d33bd4} construct more informative initializations, while DoRA~\cite{liu2024dora} and MoRA~\cite{jiang2024morahighrankupdatingparameterefficient} explore alternative update formulations. 
These methods mainly optimize the design of specific PEFT mechanisms rather than reducing redundancy within the generated trainable parameters themselves.

Another line of research investigates compact and structured parameterizations for PEFT. 
Methods such as VeRA~\cite{ICLR2024_1b53ad08}, NOLA~\cite{ICLR2024_66b99dbf}, LoRA-XS~\cite{bałazy2025loraxslowrankadaptationextremely}, SVFit~\cite{sun2024svfitparameterefficientfinetuninglarge}, and NoRA~\cite{lin2024noranestedlowrankadaptation} reduce trainable parameters through fixed bases, structured subspaces, compact cores, or nested representations. 
Tensorized and sharing-based methods further explore alternative parameter structures through decomposition or parameter sharing~\cite{yang-etal-2024-loretta,10903446,bershatsky2024lotrlowtensorrank,hounie2025lortalowranktensor,wang2026mosa,he2025rasa}. 
These existing approaches focus on introducing method-specific structures. 

In this work, we propose a novel general framework for compressing PEFT parameters via hierarchical low-rank reparameterization, called \emph{HiLoRR}. 
The core idea is to reorganize the trainable parameters of existing PEFT methods
into approximately square matrices and represent them using low-rank decomposition. 
This reorganization-decomposition operation can be applied recursively to the resulting factors, forming a hierarchical structure where parameters are compressed level by level. 
Unlike methods designed for a specific PEFT formulation, HiLoRR provides a general reparameterization framework that can be integrated with various adaptation methods, such as LoRA~\cite{hu2022lora}, DoRA~\cite{liu2024dora}, MoSA~\cite{wang2026mosa}, and RaSA~\cite{he2025rasa}.

We evaluate PEFT methods enhanced with HiLoRR on LLaMA-3.1-8B~\cite{grattafiori2024llama3herdmodels} and Qwen2.5-7B~\cite{hui2024qwen25codertechnicalreport} on commonsense reasoning benchmarks, with additional GSM8K mathematical reasoning experiments on LLaMA-3.1-8B.
Furthermore, the compressed variants reduce trainable parameters by approximately an order of magnitude while achieving higher average accuracy across the eight commonsense reasoning benchmarks than the corresponding high-parameter baselines (Table \ref{tab:commonsense_main}). 
Experiments with MoSA and RaSA further demonstrate that HiLoRR can substantially reduce trainable parameters while maintaining performance close to the original methods (Table~\ref{tab:adapter_extension}). 

The contributions of this work are summarized as follows:
\begin{itemize}
\item We introduce HiLoRR, a general hierarchical low-rank reparameterization framework for compressing trainable parameters generated by existing PEFT methods and supporting recursive application.
\item We demonstrate that HiLoRR can be applied to multiple PEFT formulations, including LoRA, DoRA, MoSA, and RaSA, without requiring method-specific redesign. 
Experiments show improved overall performance under comparable trainable-parameter budgets, as well as substantial parameter reductions with competitive performance across the evaluated PEFT methods.
\end{itemize}

\section{Related Work}

\textbf{Parameter-Efficient Fine-Tuning and LoRA.}
PEFT has become an effective paradigm for adapting large pretrained models by updating only a small subset of parameters while keeping the backbone frozen. Early PEFT
approaches mainly introduced additional trainable modules or optimized
input representations, including adapter modules, prompt tuning, and
prefix tuning~\cite{pmlr-v97-houlsby19a,lester-etal-2021-power,li-liang-2021-prefix}.
Among these approaches, Low-Rank Adaptation (LoRA)~\cite{hu2022lora} has become one of the most widely adopted methods due to its simplicity and strong empirical performance. 
LoRA represents the weight update as a low-rank decomposition, which is optimized while freezing the pretrained weights. 
This formulation substantially reduces trainable parameters while maintaining competitive
performance. 
Under the LoRA formulation, the adaptation rank simultaneously determines the trainable parameter budget and the rank upper bound of the update matrix. 
Therefore, reducing the parameter budget by decreasing the rank also restricts the adaptation space.

\textbf{LoRA Variants.}
Following the success of LoRA, numerous variants have been proposed to
improve its parameter efficiency, optimization behavior, and adaptation
capacity. Rank-adaptive methods, such as AdaLoRA~\cite{zhang2023adaptive} and DyLoRA~\cite{valipour-etal-2023-dylora}, dynamically
adjust effective ranks or parameter budgets to better utilize limited
trainable parameters. 
Optimization-oriented approaches, including LoRA+~\cite{hayou2024loraefficientlowrank} and rsLoRA~\cite{kalajdzievski2023rankstabilizationscalingfactor}, improve
the training dynamics of LoRA factors through asymmetric learning rates
and improved scaling strategies.
Other methods focus on initialization and update capacity. PiSSA~\cite{NEURIPS2024_db36f4d6}, EVA~\cite{NEURIPS2025_41d33bd4},
and LoRA-GA~\cite{NEURIPS2024_62c4718c} exploit pretrained weights, activation statistics, or
gradient information to construct more informative initializations. 
DoRA~\cite{liu2024dora} decomposes weights into magnitude and direction components and uses LoRA for directional updates. 
ReLoRA~\cite{lialin2023stack} periodically merges and reinitializes low-rank updates, whereas
MoRA~\cite{jiang2024morahighrankupdatingparameterefficient} uses a square trainable matrix with non-parametric operators to construct higher-rank updates.
These methods mainly improve how LoRA ranks are allocated, how its factors are optimized or initialized, or how the resulting updates are constructed. 
Different from them, we focus on recursively using the low-rank parameterization to reorganize their trainable parameters.

\textbf{Compact and Structured Parameterizations.}
A complementary line of research focuses on reducing task-specific parameters through compact representations and structured reparameterization. 
VeRA~\cite{ICLR2024_1b53ad08} shares frozen random matrices across layers and learns only scaling vectors, while Tied-LoRA~\cite{renduchintala-etal-2024-tied} exploits parameter
sharing and LoRA-FA~\cite{zhang2026lorafaefficienteffectivelow} freezes one LoRA factor to reduce redundancy or memory consumption.
NOLA~\cite{ICLR2024_66b99dbf} represents LoRA factors using combinations of fixed random
matrices, whereas LoRA-XS~\cite{bałazy2025loraxslowrankadaptationextremely} uses SVD-derived bases from pretrained weights and learns only a compact core.
SVFit~\cite{sun2024svfitparameterefficientfinetuninglarge} further exploits the singular structure of pretrained weights by freezing singular-vector subspaces and optimizing only the most significant singular values.
Nested Low-Rank Adaptation (NoRA)~\cite{lin2024noranestedlowrankadaptation} introduces a nested structure by freezing an outer low-rank representation and optimizing an inner
low-rank adapter.
Beyond factor-level compression, tensorized approaches such as LoRETTA~\cite{yang-etal-2024-loretta}, TT-LoRA~\cite{10903446}, LoTR~\cite{bershatsky2024lotrlowtensorrank}, and LoRTA~\cite{hounie2025lortalowranktensor} utilize higher-order decompositions to construct
compact adaptation representations.
Kronecker-based methods and folding-based approaches further explore structured update representations~\cite{Edalati2025,
chen2024superloraparameterefficientunifiedadaptation}.
Sharing-based methods, such as MoSA~\cite{wang2026mosa} and RaSA~\cite{he2025rasa}, reduce adaptation overhead through fine-grained parameter sharing or cross-layer rank sharing.
Unlike these methods that introduce specific structures for individual PEFT formulations, our work proposes a general hierarchical low-rank reparameterization framework that recursively reorganizes the trainable PEFT parameters into compact structures and applies low-rank decomposition level by level. The proposed framework is compatible with various PEFT methods, such as LoRA, DoRA, MoSA, and RaSA.

\begin{figure*}[t]
	\raggedright
	\definecolor{skyblue}{RGB}{200,230,242}
	\resizebox{\textwidth}{!}{%
	\begin{tikzpicture}[
	>={Stealth[length=2.2mm]},
	font=\small,
	mat/.style={draw, thick, rounded corners=1pt, minimum height=9mm,
		align=center, fill=blue!5},
	matB/.style={mat, fill=orange!12},
	matA/.style={mat, fill=teal!12},
	fold/.style={mat, fill=purple!10},
	op/.style={circle, draw, thick, inner sep=1pt, minimum size=5mm},
	lbl/.style={font=\scriptsize, text=black!70},
	flow/.style={->, thick, black!60},
	]
	
	\node (title1) [font=\small\bfseries, anchor=west] at (-8.75,0) {(a) Standard LoRA};
	\node[mat, minimum width=18mm, minimum height=16mm]  (W0) at (-4.85,-1.4) {$W_0$\\[-1pt]{\scriptsize $m{\times}n$}};
	\node[op, right=6mm of W0]      (plus) {$+$};
	\node[matB, minimum width=11mm, minimum height=16mm, right=6mm of plus] (B1) {$B$\\[-1pt]{\scriptsize $m{\times}r$}};
	\node[op, right=3mm of B1]      (dot1) {$\cdot$};
	\node[matA, minimum width=18mm, minimum height=11mm, right=3mm of dot1] (A1) {$A$\\[-1pt]{\scriptsize $r{\times}n$}};
	
	\draw[flow] (W0) -- (plus);
	\draw[flow] (plus) -- (B1);
	
	\node (title2) [font=\small\bfseries, anchor=west] at (-8.75,-2.9)
	{(b) HiLoRA: Hierarchical Low-Rank Reparameterization of LoRA};
	\node (row2anchor) at (-1,-2.9) {};
	
	\node[matB, fill=gray!15, minimum width=11mm, minimum height=16mm, below=6mm of row2anchor, xshift=-56mm] (B2)
	{$B$\\[-1pt]{\scriptsize $m{\times}r$}};
	\node[fold, fill=gray!15, minimum width=12.8mm, minimum height=12.8mm, right=36mm of B2] (M)
	{$B_\text{reshape}$\\[-1pt]{\scriptsize $p_B{\times}q_B$}};
	\node[fold, minimum width=6.4mm, minimum height=12.8mm, right=24.5mm of M] (U)
	{$U_{\scriptscriptstyle B}$\\[-1pt]{\scriptsize $p_B{\times}k_B$}};
	\node[op, right=2.5mm of U] (dotUV) {$\cdot$};
	\node[fold, minimum width=12.8mm, minimum height=6.4mm, right=2.5mm of dotUV] (V)
	{$V_{\scriptscriptstyle B}$\\[-1pt]{\scriptsize $k_B{\times}q_B$}};
	\node[matB, fill=skyblue, minimum width=11mm, minimum height=16mm, right=15mm of V] (B3)
	{$B$\\[-1pt]{\scriptsize $m{\times}r$}};
	
	\begin{scope}[on background layer]
	\node[draw, dashed, rounded corners, fill=purple!4,
	fit=(U)(dotUV)(V), inner sep=3mm] (train) {};
	\end{scope}

	\draw[flow] (B2) -- node[lbl, above]{Near-Square Reshaping} node[lbl, below]{$p_B\,q_B{=}mr$} (M);
	\draw[flow] (M)  -- node[lbl, above]{Parameterization} node[lbl, below]{rank $k_B$} (train);
	\draw[flow] (train)  -- node[lbl, above]{Back} node[lbl, below]{\scriptsize $m{\times}r$} (B3);

	\node[lbl, below=0.5mm of train, text=purple!60!black]
	{trainable: $U_B,V_B$};
	
	\node[matA, fill=gray!15, minimum width=18mm, minimum height=11mm, below=12mm of B2] (A2)
	{$A$\\[-1pt]{\scriptsize $r{\times}n$}};
	\node[fold, fill=gray!15, minimum width=13mm, minimum height=13mm, right=36mm of A2] (MA)
	{$A_\text{reshape}$\\[-1pt]{\scriptsize $p_A{\times}q_A$}};
	\node[fold, minimum width=6.5mm, minimum height=13mm, right=24.0mm of MA] (UA)
	{$U_{\scriptscriptstyle A}$\\[-1pt]{\scriptsize $p_A{\times}k_A$}};
	\node[op, right=2.5mm of UA] (dotUVA) {$\cdot$};
	\node[fold, minimum width=13mm, minimum height=6.5mm, right=2.5mm of dotUVA] (VA)
	{$V_{\scriptscriptstyle A}$\\[-1pt]{\scriptsize $k_A{\times}q_A$}};
	\node[matA, fill=skyblue, minimum width=18mm, minimum height=11mm, right=15mm of VA] (A3)
	{$A$\\[-1pt]{\scriptsize $r{\times}n$}};
	
	\begin{scope}[on background layer]
	\node[draw, dashed, rounded corners, fill=purple!4,
	fit=(UA)(dotUVA)(VA), inner sep=3mm] (trainA) {};
	\end{scope}

	\draw[flow] (A2) -- node[lbl, above]{Near-Square Reshaping} node[lbl, below]{$p_A\,q_A{=}rn$} (MA);
	\draw[flow] (MA) -- node[lbl, above]{Parameterization} node[lbl, below]{rank $k_A$} (trainA);
	\draw[flow] (trainA) -- node[lbl, above]{Back} node[lbl, below]{{\scriptsize $r{\times}n$}} (A3);

	\node[lbl, below=0.5mm of trainA, text=purple!60!black]
	{trainable: $U_A,V_A$};
	
	\end{tikzpicture}}
    
	\caption{\textbf{Overview of HiLoRA (our method HiLoRR applied to LoRA).}
		(a) Standard LoRA adds a rank-$r$ update $\Delta W=\frac{\alpha}{r}BA$ to the frozen weight $W_0$. 
		(b) HiLoRA reshapes each LoRA factor into a Near-Square matrix ($B_\text{reshape}$/$A_\text{reshape}$) and represents it by a second low-rank product $U_BV_B$ (resp.\ $U_AV_A$). 
		The factor is then reshaped back into the reconstructed $B$/$A$ (blue boxes), 
        so the maximum rank of the matrix product $BA$ remains bounded by $r$.
		%
		%
		HiLoRA directly parameterizes and trains only $U_B,V_B,U_A,V_A$, from which the factors ($B$/$A$ (blue boxes)) are reconstructed on the fly.}
	\label{fig:method}
\end{figure*}
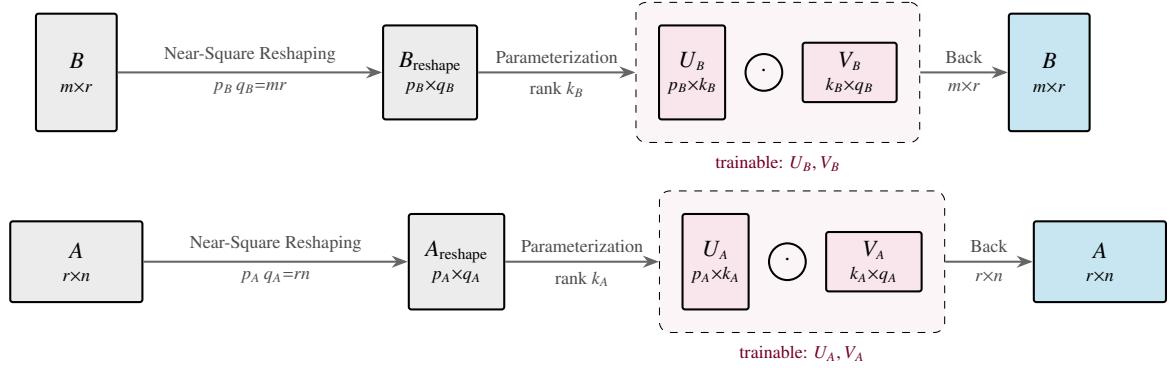

\section{Method}\label{method}

\subsection{Overview of HiLoRR}

HiLoRR is a hierarchical low-rank reparameterization framework for PEFT by reorganizing trainable PEFT parameters into approximately square matrices that are represented via low-rank decompositions, and recursively running the reorganization and decomposition. 

Unlike existing approaches to designing method-specific parameter structures, HiLoRR operates directly on the trainable parameters of existing PEFT formulations.
Given a PEFT method with a trainable parameter set $\Theta=\{\theta_1,\theta_2,\ldots,\theta_N\}$, HiLoRR aims to reduce the redundancy inside $\Theta$ through hierarchical low-rank reparameterization. 

For each trainable parameter tensor $\theta_i$, we first reshape it into an approximately square matrix:
\begin{equation}
M_i=\mathrm{reshape}(\theta_i) \in \mathbb{R}^{p_i \times q_i},
\end{equation}
where the reshaping operation preserves the number of elements $|\theta_i|$, i.e., 
satisfying $p_i q_i = |\theta_i|$ with $p_i\approx q_i$.

The reshaped matrix is then represented by a low-rank decomposition:
\begin{equation}
M_i= U_i^{(1)}V_i^{(1)},
\end{equation}
where $U_i^{(1)} \in \mathbb{R}^{p_i \times k_i}$ and $V_i^{(1)} \in \mathbb{R}^{k_i \times q_i}$ are two low-rank matrices (i.e., $0 < k_i \ll p_i$ and $0 < k_i \ll q_i$) that together contain $k_i(p_i+q_i)$ trainable parameters. 

Theoretically, HiLoRR can recursively apply the reshape-decomposition procedure to $U_i^{(1)}$ and $V_i^{(1)}$, thereby further reducing the number of trainable parameters.
Finally, this recursive process forms a hierarchical structure.
During training, only the low-rank matrices at the deepest level are optimized, and the original PEFT parameter tensor is reconstructed via the inverse process of the reparameterization. 

\subsection{Applying HiLoRR to LoRA}
We first discuss LoRA as an example PEFT instantiation. Fig.~\ref{fig:method} illustrates how HiLoRR is applied to LoRA.

\paragraph{Preliminaries of LoRA}\label{sec:prelim}
Given a frozen linear weight $W_0\in\mathbb{R}^{m\times n}$, LoRA adds a low-rank update and keeps $W_0$ fixed:
\begin{equation}
W = W_0 + \Delta W, \qquad
\Delta W = \frac{\alpha}{r}\,B A ,
\label{eq:lora}
\end{equation}
where $B\in\mathbb{R}^{m\times r}$, $A\in\mathbb{R}^{r\times n}$, $r\ll\min(m,n)$. Here, $r$ denotes the \textbf{outer rank}, which determines the maximum rank of the final weight update $\Delta W$. 
Since $\operatorname{rank}(\Delta W)\leq r$, the outer rank $r$ controls the width of the adaptation pathway.
The pretrained weight $W_0$ remains frozen, and only $A$ and $B$ are optimized. 
%

%
%

\paragraph{Single-level Low-Rank Reparameterization}\label{sec:hlr}
%
For a factor $M$ with $d$ elements (e.g., $M=B$ and $d=m\cdot r$), we reshape $M$ into a matrix $M_\text{reshape} \in \mathbb{R}^{p\times q}$ with $p\,q=d$, and parameterize the reshaped matrix as a rank-$k$ product:
\begin{equation}
M_\text{reshape} = U V, \,\,
U\in\mathbb{R}^{p\times k},\;
V\in\mathbb{R}^{k\times q}.
\label{eq:fold}
\end{equation}
Here, $k$ is called the \textbf{hierarchical rank}, which controls the compression ratio of the reparameterization.
Different from the outer rank $r$, $k$ only determines the internal representation capacity of the reparameterized factors and does not directly change the rank
constraint of the final update $\Delta W$.

By defining an operation $\text{Back}(\cdot)$ that converts the reshaped matrix $M_\text{reshape}$ back to the original matrix $M$ (i.e., $M =\text{Back}(M_\text{reshape}) = \text{Back}(U V)$), we have:
\begin{equation}
\begin{gathered}
\Delta W = \frac{\alpha}{r}\, B A, \\
B = \text{Back}(B_\text{reshape}) = \text{Back}(U_B V_B),\\
A = \text{Back}(A_\text{reshape}) = \text{Back}(U_A V_A),
\end{gathered}
\label{eq:ffupdate}
\end{equation}
where $U_B, V_B$ (or $U_A, V_A$) form the low-rank decomposition of the reshaped matrix $B_\text{reshape}$ (or $A_\text{reshape}$).
In practice, the trainable parameters are now $U$ and $V$ instead of the original parameter tensor $M$, and the number of trainable parameters changes from $d$ to $(p+q)\times k$.
If we find $k$ satisfying $ k < d/(p+q) = pq/(p+q)$, the trainable parameters are compressed.
%
The reshape and $\text{Back}(\cdot)$ operations are pure index rearrangements without arithmetic transformation. 
%



\paragraph{Near-Square Reshaping}\label{sec:nearsquare}
The shape $(p,q)$ is a design choice. 
HiLoRA (HiLoRR applied to LoRA) uses \emph{Near-Square} reshaping to determine the matrix dimensions $(p,q)$. This choice is motivated by SuperLoRA~\cite{chen2024superloraparameterefficientunifiedadaptation}, which applied square-like reshaping to grouped weight updates and reported better performance than its corresponding unreshaped form in low-parameter vision-transfer settings. While SuperLoRA reshapes grouped weight updates, HiLoRA reshapes each trainable LoRA factor separately.
%

Specifically, among all positive integer pairs satisfying $pq=d$, we pick the divisor pair closest to square,
\begin{equation}
(p^{*},q^{*})=
\arg\min_{\substack{pq=d\\p,q\in\mathbb{N}}}(p+q),
\qquad p^{*}\leq q^{*}.
\label{eq:near-square}
\end{equation}
In practice, we start from $\lfloor\sqrt{d}\rfloor$ and search downward for the largest divisor $p^{*}$ of $d$, after which $q^{*}=d/p^{*}$ (so $p^{*}\le q^{*}$). 
How close $p^{*},q^{*}$ get to $\sqrt{d}$ depends on the divisors of $d$: 
when $d$ admits a divisor near $\sqrt{d}$, the reshaping is nearly square ($p^{*}\approx q^{*}\approx\sqrt{d}$), whereas a $d$ with only unbalanced divisors yields an elongated reshaping with
$p^{*}+q^{*}\gg 2\sqrt{d}$. 
Transformer factor sizes are typically highly composite, so the Near-Square case is common.

%
%
%
%
%

\paragraph{Hierarchical Reparameterization}
This low-rank decomposition can be applied recursively to obtain a hierarchical reparameterization. 
For example, $U$ in Eq.~\eqref{eq:fold} can be further reshaped and then decomposed as a product of two low-rank matrices. 
Using the hierarchical reparameterization, the learnable parameters are the matrices at the deepest level.
%

\subsection{Applying HiLoRR to Different PEFT Methods}
HiLoRR is a general reparameterization framework that can be integrated into various PEFT methods.

\textbf{HiLoRR for LoRA-based Methods.}
For PEFT methods derived from LoRA, HiLoRR is applied to the trainable components of the original formulation while preserving their specific design principles.

For DoRA, which decomposes the pretrained weight into magnitude and direction components, HiLoRR is applied to the trainable directional parameters. 
Specifically, the directional component is replaced by its hierarchical low-rank representation, while the magnitude component is kept in its original form. 
The resulting method, referred to as \emph{HiDoRA}, inherits the magnitude-direction decomposition of DoRA while further reducing the number of trainable parameters through hierarchical
reparameterization.

Similarly, RaSA introduces rank sharing across different layers to improve parameter efficiency. HiLoRR can be incorporated into RaSA by applying hierarchical reparameterization to the trainable rank-sharing parameters while preserving the original cross-layer sharing mechanism.
The resulting method, denoted as \emph{HiRaSA}, combines the sharing capability of RaSA with the additional compression capability of HiLoRR.

\textbf{HiLoRR for General PEFT Parameterizations.}
For PEFT methods whose trainable parameters do not follow the LoRA-style factorized formulation, HiLoRR directly operates on the generated trainable parameters.

For example, MoSA represents its trainable parameters as shared parameter vectors. HiLoRR first reorganizes each trainable parameter vector into an approximately square matrix, then applies a low-rank decomposition.
The same hierarchical decomposition process can be recursively applied to the resulting factors. 
The resulting method, referred to as \emph{HiMoSA}, maintains the original parameter sharing mechanism of MoSA while further compressing its trainable parameters through
hierarchical low-rank reparameterization.


%

\subsection{Optimization Characteristics of Hierarchical Reparameterization}
\label{sec:ff-optimization}
The hierarchical factorization changes not only the number of trainable parameters, but also the optimization behavior.

Here, we take the single-level low-rank reparameterization for LoRA as an example for discussion.
In the original LoRA formulation, let $c=\alpha/r$ and $G=\partial\mathcal{L}/\partial\Delta W$ denote the gradient with respect to the induced weight update. 
The gradients of the two LoRA factors are
\begin{equation}
\nabla_B \mathcal{L}
=
c\,G A^{\top},
\qquad
\nabla_A \mathcal{L}
=
c\,B^{\top}G.
\label{eq:lora-gradient}
\end{equation}
Thus, the gradient of each trainable factor is modulated by only the
other LoRA factor.

%
%
For HiLoRA, since the operation $\text{Back}(\cdot)$ is a permutation of entries, its backward operator
$\mathcal{R}^{*}$ simply restores the gradient to the corresponding reshaped shape. The gradients of the four trainable factors are
\begin{align}
\nabla_{U_B}\mathcal{L}
&=
\mathcal{R}_B^{*}
\left(c\,GA^{\top}\right)V_B^{\top},
&
\nabla_{V_B}\mathcal{L}
&=
U_B^{\top}
\mathcal{R}_B^{*}
\left(c\,GA^{\top}\right),
\label{eq:ff-gradient-b}
\\
\nabla_{U_A}\mathcal{L}
&=
\mathcal{R}_A^{*}
\left(c\,B^{\top}G\right)V_A^{\top},
&
\nabla_{V_A}\mathcal{L}
&=
U_A^{\top}
\mathcal{R}_A^{*}
\left(c\,B^{\top}G\right).
\label{eq:ff-gradient-a}
\end{align}
%
The companion factors $A$ and $B$ are themselves reconstructed from two trainable matrices. 
Therefore,
the gradient of each internal parameter is modulated by three trainable factors, up to the norm-preserving reshape operations.
For example,
\begin{equation}
\left\|
\nabla_{U_B}\mathcal{L}
\right\|_F
\leq
c\,
\|G\|_F
\|U_A\|_F
\|V_A\|_2
\|V_B\|_2,
\label{eq:ff-gradient-bound}
\end{equation}
with analogous bounds for the other internal factors. In contrast,
the corresponding standard-LoRA gradient in
Eq.~\eqref{eq:lora-gradient} depends on only one trainable companion
factor. Consequently, the effective gradient scale in HiLoRA is more
sensitive to the initialization and norms of its internal factors.
When these factors are initialized at small or moderate scales, the
deeper multiplicative path can attenuate the early optimization
information.

This effect is further amplified by the LoRA-compatible zero
initialization used in HiLoRA. For example, when $U_B=0$ at
initialization while $V_B$, $U_A$, and $V_A$ are nonzero, we have
$B=0$ and
\begin{equation}
\nabla_{V_B}\mathcal{L}
=
\nabla_{U_A}\mathcal{L}
=
\nabla_{V_A}\mathcal{L}
=
0,
\label{eq:ff-zero-init-gradient}
\end{equation}
whereas $\nabla_{U_B}\mathcal{L}$ is generally nonzero. Hence, only one of the four internal factors is updated at the first optimization step; the remaining factors receive nonzero gradients only after $U_B$ moves away from zero. 
This progressive activation preserves the initial condition $\Delta W=0$, but can make the early-stage optimization of HiLoRA slower than that of the original PEFT formulation.

These observations motivate a larger learning rate for the reshaped matrices to compensate for their reduced and delayed effective update scale. We therefore treat the learning rate of HiLoRA as a method-specific hyperparameter rather than directly inheriting the default value used by LoRA. 
Importantly, the analysis does not imply a universal analytical multiplier: the required compensation depends on the reshaped rank, factor initialization, scaling coefficient, and optimizer. 
Our ablation experiments empirically verify that HiLoRA generally achieves its best performance at a higher learning rate than the original PEFT formulation.
\section{Experiments}\label{sec:results}

We evaluate HiLoRR as a general hierarchical low-rank reparameterization
framework on two model families and two reasoning tasks. Our experiments
are organized around three questions: (i) can HiLoRA/HiDoRA reduce trainable
parameters while preserving the performance of LoRA/DoRA on reasoning tasks?
(ii) can the proposed reparameterization be extended to different PEFT
formulations? and (iii) how do different hierarchical configurations
affect the compression-performance trade-off?

\subsection{Experimental Setup}
\label{sec:setup}

\paragraph{Models}
We use two open base (non-instruct) models, i.e.,
\textbf{LLaMA-3.1-8B}~\cite{grattafiori2024llama3herdmodels} and \textbf{Qwen2.5-7B}~\cite{hui2024qwen25codertechnicalreport}.
Base models are chosen so that the reported gains reflect what the PEFT method itself learns, rather than instruction-tuning already baked into the backbone. 
The two models differ in architecture (grouped-query attention configuration, intermediate width), allowing us to test cross-architecture generalization.

\begin{table*}[!t]
	\caption{Accuracy (\%) on eight commonsense reasoning benchmarks using LLaMA-3.1-8B and Qwen2.5-7B. Within each model, methods are grouped by trainable-parameter budget. Mem is training-phase peak PyTorch CUDA allocated memory in GiB. Rows shaded in gray indicate our methods.}
	\label{tab:commonsense_main}
	\centering
	\footnotesize
	\setlength{\tabcolsep}{1.5pt}
	\renewcommand{\arraystretch}{0.96}
	\setlength{\aboverulesep}{1.2pt}
	\setlength{\belowrulesep}{1.2pt}
	\begin{tabularx}{0.90\textwidth}{@{}lcc*{10}{>{\centering\arraybackslash}X}@{}}
		\toprule
		Method & \shortstack{Params(M)} & \shortstack{Mem(GiB)} & BoolQ & PIQA & SIQA & ARC-c & ARC-e & OBQA & HS & WG & Avg. & W-Avg. \\
		\midrule
		\multicolumn{13}{@{}l}{\textbf{LLaMA-3.1-8B}} \\
		\addlinespace[1pt]
		\multicolumn{13}{@{}l}{\quad\emph{Low-parameter ($\leq 6$M)}} \\
		LoRA-r2     & 3.54 & 11.149 & 62.17 & 86.40 & 75.59 & 80.46 & 91.67 & 82.80 & 90.26 & 74.11 & 80.43 & 83.13 \\
		DoRA-r2     & 4.33 & 11.160 & 62.17 & 86.45 & 75.38 & 81.40 & 91.54 & 84.00 & 90.40 & 74.27 & 80.70 & 83.25 \\
		SVFit       & 0.39 & 27.346 & 67.86 & 84.11 & 68.37 & 77.30 & 90.87 & 75.40 & 82.32 & 60.54 & 75.85 & 78.40 \\
		NoRA        & 5.24 & 24.675 & 62.17 & 86.51 & 75.59 & 82.00 & 92.68 & 82.40 & 91.20 & 76.80 & 81.17 & 83.88 \\
		\rowcolor{black!8} HiLoRA    & 2.82 & 11.138 & 71.04 & 87.11 & 78.20 & 81.31 & 91.62 & 84.40 & 91.74 & 80.82 & 83.28 & 85.82 \\
		\rowcolor{black!8} HiDoRA    & 3.60 & 11.150 & 62.17 & 86.72 & 77.64 & 81.31 & 92.42 & 86.00 & 91.69 & 80.51 & 82.31 & 84.53 \\
		\addlinespace[1pt]
		\multicolumn{13}{@{}l}{\quad\emph{High-parameter ($\geq 26$M)}} \\
		LoRA-r16    & 28.31 & 11.518 & 62.17 & 86.94 & 76.25 & 81.40 & 92.55 & 85.00 & 92.03 & 77.35 & 81.71 & 84.39 \\
		DoRA-r16    & 29.10 & 11.529 & 62.17 & 86.24 & 76.92 & 83.19 & 92.97 & 83.40 & 91.89 & 78.14 & 81.86 & 84.48 \\
		MoRA        & 28.23 & 11.545 & 62.17 & 84.87 & 78.25 & 80.12 & 91.58 & 84.20 & 91.83 & 81.29 & 81.79 & 84.35 \\
		\rowcolor{black!8} HiLoRA    & 28.18 & 11.517 & 69.76 & 87.76 & 78.30 & 82.76 & 92.17 & 85.60 & 92.47 & 80.98 & 83.73 & 86.19 \\
		\rowcolor{black!8} HiDoRA    & 28.44 & 11.527 & 71.35 & 87.32 & 79.27 & 82.51 & 91.75 & 85.40 & 92.95 & 81.37 & 83.99 & 86.65 \\
		\midrule
		\multicolumn{13}{@{}l}{\textbf{Qwen2.5-7B}} \\
		\addlinespace[1pt]
		\multicolumn{13}{@{}l}{\quad\emph{Low-parameter ($\leq 6$M)}} \\
		LoRA-r2     & 3.38 & 11.336 & 65.26 & 88.25 & 79.27 & 89.51 & 95.96 & 89.40 & 92.35 & 76.95 & 84.62 & 86.22 \\
		DoRA-r2     & 4.14 & 11.689 & 66.15 & 88.36 & 79.63 & 88.99 & 95.88 & 89.80 & 92.61 & 76.95 & 84.80 & 86.48 \\
		SVFit       & 0.29 & 25.699 & 69.05 & 85.85 & 74.62 & 86.26 & 95.62 & 86.00 & 83.56 & 68.51 & 81.18 & 81.48 \\
		NoRA        & 4.59 & 22.950 & 65.96 & 89.06 & 79.38 & 88.65 & 95.54 & 91.00 & 93.17 & 78.61 & 85.17 & 86.81 \\
		\rowcolor{black!8} HiLoRA    & 2.79 & 11.326 & 70.24 & 89.12 & 80.25 & 88.99 & 95.83 & 90.40 & 93.30 & 79.40 & 85.94 & 87.65 \\
		\rowcolor{black!8} HiDoRA    & 3.55 & 11.338 & 69.91 & 89.39 & 79.73 & 88.74 & 95.79 & 91.40 & 93.06 & 79.64 & 85.96 & 87.49 \\
		\addlinespace[1pt]
		\multicolumn{13}{@{}l}{\quad\emph{High-parameter ($\geq 26$M)}} \\
		LoRA-r16    & 27.07 & 11.702 & 70.55 & 89.28 & 79.12 & 88.91 & 95.75 & 90.40 & 92.95 & 79.56 & 85.81 & 87.45 \\
		DoRA-r16    & 27.83 & 12.061 & 70.15 & 89.39 & 79.38 & 88.99 & 95.96 & 90.20 & 92.81 & 79.72 & 85.82 & 87.39 \\
		MoRA        & 27.03 & 11.707 & 62.17 & 88.41 & 79.22 & 88.31 & 95.33 & 90.40 & 92.93 & 81.06 & 84.73 & 86.16 \\
		\rowcolor{black!8} HiLoRA    & 26.64 & 11.682 & 72.26 & 89.34 & 80.09 & 88.74 & 95.66 & 91.20 & 93.18 & 80.66 & 86.39 & 87.95 \\
		\rowcolor{black!8} HiDoRA    & 27.73 & 11.698 & 72.05 & 89.50 & 80.30 & 89.51 & 95.71 & 91.40 & 92.99 & 80.51 & 86.49 & 87.91 \\
		\bottomrule
	\end{tabularx}
	\par\smallskip
	\begin{minipage}{\textwidth}\scriptsize\raggedright
	\end{minipage}
\end{table*}

\paragraph{Tasks and Datasets}

For commonsense reasoning, we fine-tune the models on the Commonsense-15K instruction dataset~\cite{NEURIPS2024_48c368f1}. After fine-tuning, the models are evaluated on eight widely used commonsense reasoning benchmarks,
including BoolQ~\cite{clark-etal-2019-boolq}, PIQA~\cite{bisk2020piqa}, SIQA~\cite{sap-etal-2019-social}, HellaSwag~\cite{zellers-etal-2019-hellaswag}, WinoGrande~\cite{10.1145/3474381}, ARC-c, ARC-e~\cite{clark2018thinksolvedquestionanswering}, and
OBQA~\cite{mihaylov-etal-2018-suit} in Table~\ref{tab:commonsense_main}.
We evaluate each benchmark separately and report the accuracy (\%) on each task. 
In addition, 
we report two overall performance metrics across the eight tasks: the unweighted macro average (Avg.) and the weighted average (W-Avg.), with weights proportional to the number of test examples in each task.

For mathematical reasoning, we fine-tune LLaMA-3.1-8B on the MetaMathQA-40K~\cite{ICLR2024_c400474e} instruction dataset and evaluate it on the GSM8K benchmark~\cite{cobbe2021trainingverifierssolvemath}. We report the exact-match
accuracy of the generated final answers on the official GSM8K test set in Table~\ref{tab:gsm8k_main}.

In both tables, fine-tuned methods are grouped into low-parameter
($\leq 6$M) and high-parameter ($\geq 26$M) settings. We also report
peak GPU memory (Mem) for both commonsense
reasoning and mathematical reasoning to show
training-resource demands alongside accuracy and trainable-parameter
count. Here, Mem denotes the peak PyTorch CUDA allocated memory during training.


\paragraph{Baselines and Targeted Modules}

We compare HiLoRA and HiDoRA with representative PEFT methods, including LoRA, DoRA, MoRA, NoRA, and SVFit. All methods are applied to the same target modules, including the self-attention projection matrices $W_Q$, $W_K$ and $W_V$, as well as the FFN projection matrices $W_{\mathrm{up}}$ and $W_{\mathrm{down}}$ in each Transformer block. All remaining parameters of the pretrained backbone are kept frozen during training.




\paragraph{Parameter-Budget Settings}

For fair comparisons, we control the trainable parameter budget. 
The number of trainable parameters introduced by LoRA with rank
$r=16$ serves as the reference parameter budget.
For compressed settings, HiLoRA is configured to use approximately $10\%$
of the trainable parameters of the original LoRA baseline while
maintaining the same target modules. 
For matched-parameter settings, we jointly increase the outer rank
$r$ and hierarchical rank $k$ of HiLoRA until the total number of
trainable parameters is approximately equal to that of the original LoRA baseline.
The parameter counts reported in the tables correspond to the number of trainable parameters introduced by each PEFT method. {The detailed discussion of parameter compression is provided in Appendix~\ref{sec:paramanalysis}.}

\paragraph{Optimization and Schedules}

We optimize all methods using AdamW~\cite{loshchilov2018decoupled} with a cosine learning-rate schedule and a warm-up ratio of 0.1. 
The candidate learning rates are selected from ranges commonly used in PEFT experiments. 
We observe that HiLoRA generally benefits from a larger peak learning rate (Table~\ref{tab:lr_ablation}).
%
%
For HiLoRA and HiDoRA, following our optimization analysis, we adopt a larger learning rate of $2\times10^{-3}$ due to the additional hierarchical factorization structure. 

\subsection{Results: Commonsense Reasoning}
\label{sec:comparison}
We evaluate HiLoRA and HiDoRA on the commonsense reasoning tasks. 
Table~\ref{tab:commonsense_main} presents results on LLaMA-3.1-8B and Qwen2.5-7B under low- and
high-parameter settings. 
In each setting, the highest Avg. and W-Avg. scores are obtained by either HiLoRA or HiDoRA.  
\textbf{\emph{In the low-parameter setting}}, both methods achieve higher Avg. and W-Avg. scores than all baselines reported in Table~\ref{tab:commonsense_main} on both backbones, while using approximately 10--13$\%$ of the trainable parameters of the corresponding LoRA-r16 baseline.
These results show that substantial parameter compression can maintain strong average accuracy across the eight benchmarks, while directly reducing the LoRA rank to decrease the number of trainable parameters can lead to degraded task performance.
\textbf{\emph{In the high-parameter setting}}, HiLoRA and HiDoRA outperform LoRA and DoRA on both aggregate metrics with comparable numbers of trainable parameters. This indicates that hierarchical reparameterization can also improve performance at a similar parameter budget.
\textbf{\emph{In terms of training memory}}, HiLoRA and HiDoRA have peak allocated GPU memory comparable to LoRA and DoRA and substantially lower than the reported NoRA and SVFit configurations.

\subsection{Results: Mathematical Reasoning}
We evaluate HiLoRA and HiDoRA on the GSM8K mathematical reasoning benchmark. 
Table~\ref{tab:gsm8k_main} reports the results on LLaMA-3.1-8B under low- and high-parameter settings.
For reference, we also report the performance of the backbone without task-specific supervised fine-tuning, denoted as Base (no SFT).
\textbf{\emph{In the low-parameter setting,}} HiLoRA and HiDoRA achieve the two highest EM scores in this group, at 70.20\% and 70.05\%, respectively, although both remain below the high-parameter LoRA-r16 baseline.
\textbf{\emph{In the high-parameter setting,}} HiLoRA and HiDoRA achieve EM scores of 73.16\% and 73.24\%, respectively, outperforming all compared baselines with comparable numbers of trainable parameters.
\textbf{\emph{In terms of training memory,}} both methods show the same pattern as in commonsense reasoning.
These results show that hierarchical low-rank reparameterization achieves competitive GSM8K performance with substantially fewer trainable parameters and improves EM scores at a comparable parameter budget.

\begin{table}[!t]
	\caption{Accuracy (\%) on GSM8K after MetaMathQA-40K fine-tuning using LLaMA-3.1-8B. Our methods are shaded in gray.}
	\label{tab:gsm8k_main}
	\centering
	\footnotesize
	\setlength{\tabcolsep}{1.5pt}
	\renewcommand{\arraystretch}{0.96}
	\setlength{\aboverulesep}{1.2pt}
	\setlength{\belowrulesep}{1.2pt}
	\renewcommand{\tabularxcolumn}[1]{m{#1}}
	\begin{tabularx}{\columnwidth}{@{}lcc*{2}{>{\centering\arraybackslash}X}@{}}
		\toprule
		Method & \shortstack{Params\\(M)} & \shortstack{Mem\\(GiB)} & \shortstack{Correct\\(/1,319)} & \shortstack{GSM8K\\EM(\%)} \\
		\specialrule{\lightrulewidth}{1.2pt}{2.0pt}
        Base (no SFT) & 0.00 & -- & 390 & 29.57 \\
        \specialrule{\lightrulewidth}{0.4pt}{1.2pt}
		\multicolumn{5}{@{}l}{\emph{Low-parameter ($\leq 6$M)}} \\
		LoRA-r2 & 3.54 & 15.836 & 917 & 69.52 \\
		DoRA-r2 & 4.33 & 15.844 & 890 & 67.48 \\
		SVFit & 0.39 & 31.447 & 711 & 53.90 \\
		NoRA & 5.24 & 29.356 & 889 & 67.40 \\
		\rowcolor{black!8} HiLoRA & 2.82 & 15.828 & 926 & 70.20 \\
		\rowcolor{black!8} HiDoRA & 3.60 & 15.836 & 924 & 70.05 \\
		\midrule
		\multicolumn{5}{@{}l}{\emph{High-parameter ($\geq 26$M)}} \\
		LoRA-r16 & 28.31 & 16.113 & 952 & 72.18 \\
		DoRA-r16 & 29.10 & 16.121 & 957 & 72.55 \\
		MoRA & 28.23 & 16.131 & 951 & 72.10 \\
		\rowcolor{black!8} HiLoRA & 28.18 & 16.112 & 965 & 73.16 \\
		\rowcolor{black!8} HiDoRA & 28.44 & 16.114 & 966 & 73.24 \\
		\bottomrule
	\end{tabularx}
	\par\smallskip
	\begin{minipage}{\columnwidth}\scriptsize\raggedright
	\end{minipage}
\end{table}




\subsection{Results: Generalization Across PEFT Methods}
{To evaluate the applicability of HiLoRR across PEFT methods, we apply it to two additional methods, MoSA and RaSA, obtaining HiMoSA and HiRaSA, respectively, while preserving their original adaptation mechanisms. 
Table~\ref{tab:adapter_extension}  reports their results on the eight commonsense reasoning benchmarks using LLaMA-3.1-8B.
HiMoSA and HiRaSA reduce trainable parameters by 92.6$\%$ and 86.9$\%$, respectively, while maintaining average accuracy close to their original counterparts. These results support the applicability of HiLoRR beyond LoRA and DoRA.}

\begin{table*}[!t]
\caption{Accuracy (\%) on eight commonsense reasoning benchmarks when applying hierarchical low-rank reparameterization to other adapter families. All results use LLaMA-3.1-8B. }
\label{tab:adapter_extension}
\centering
\scriptsize
\setlength{\tabcolsep}{3.0pt}
\scalebox{1.09}{
\begin{tabular}{lccccccccccc}
\toprule
Method & Params (M) & BoolQ & PIQA & SIQA & ARC-c & ARC-e & OBQA & HellaSwag & WinoGrande & Avg. & W-Avg. \\
\midrule
LoRA & 28.31 & 62.17 & 86.94 & 76.25 & 81.40 & 92.55 & 85.00 & 92.03 & 77.35 & 81.71 & 84.39 \\
\midrule
MoSA & 28.31 & 70.64 & 85.47 & 77.79 & 82.51 & 90.87 & 84.40 & 90.92 & 81.85 & 83.06 & 85.27 \\
\rowcolor{black!8} HiMoSA  & 2.10 & 69.54 & 86.40 & 78.25 & 81.48 & 91.96 & 83.60 & 91.20 & 81.29 & 82.97 & 85.36 \\
\midrule
RaSA & 28.32 & 69.60 & 86.18 & 79.73 & 81.06 & 92.55 & 84.20 & 92.72 & 81.14 & 83.40 & 86.21 \\
\rowcolor{black!8} HiRaSA  & 3.71 & 69.88 & 86.72 & 78.30 & 82.17 & 92.09 & 84.60 & 92.17 & 80.82 & 83.34 & 85.92 \\
\bottomrule
\end{tabular}
}
\end{table*}

\subsection{Ablation Studies}
\label{sec:ablation}

We conduct ablation studies to investigate the influence of different
HiLoRA design choices, including learning rate, reparameterization scope, reshaping
strategy, and rank configuration.

\paragraph{Learning Rate}
We investigate the sensitivity of HiLoRA to the learning rate on LLaMA-3.1-8B
by sweeping a range of values for HiLoRA.
As shown in Table~\ref{tab:lr_ablation}, HiLoRA generally achieves its
best performance at a higher learning rate. We therefore adopt a peak learning rate of $2\times10^{-3}$
for our method, as it achieves the highest Avg. score among
the evaluated learning rates.
This observation is consistent with the aforementioned optimization analysis: the additional multiplicative factorization changes the effective gradient scale, making HiLoRA benefit from a larger method-specific learning rate.

%


\begin{table*}[!t]
\caption{Learning-rate ablation of low-parameter HiLoRA on LLaMA-3.1-8B across eight commonsense reasoning tasks. Best observed values are bold.}
\label{tab:lr_ablation}
\centering
\small
\setlength{\tabcolsep}{3pt}
\renewcommand{\arraystretch}{1.16}
\begin{tabular}{@{}c*{10}{c}@{}}
\toprule
& \multicolumn{10}{c}{Learning rate ($\times 10^{-4}$)} \\
\cmidrule(l){2-11}
Metric & 1 & 3 & 5 & 8 & 10 & 12 & 16 & 20 & 24 & 32 \\
\midrule
Avg. (\%) & 74.66 & 80.09 & 80.88 & 81.65 & 82.42 & 82.21 & 83.16 & \textbf{83.28} & 81.76 & 79.52 \\
W-Avg. (\%) & 77.11 & 82.36 & 83.51 & 84.10 & 85.18 & 84.24 & \textbf{85.86} & 85.82 & 84.57 & 82.24 \\
\bottomrule
\end{tabular}
\vspace{2pt}
\end{table*}

\paragraph{Reparameterization Scope and Depth}

{We study the scope and depth of reparameterization in HiLoRA.
As shown in Table~\ref{tab:hilora_variants_ablation}, the reported
A+B variant outperforms the A-only and B-only variants at similar
trainable-parameter budgets, achieving 83.28 Avg. and 85.82 W-Avg.
We further examine the effect of reparameterization depth in
Table~\ref{tab:hilora_multilevel_budget}.
In single-level HiLoRA, each reshaped LoRA matrix, $A$ or $B$,
is parameterized as a rank-$k_1$ product $UV$.
The double-level variant recursively applies the same Near-Square
reshaping and low-rank parameterization to both $U$ and $V$,
using rank $k_2$, so that only the resulting second-level factors
are directly optimized.
We fix the outer LoRA rank at $r=16$ and adjust the internal ranks
to compare single- and double-level variants at similar low,
medium, and high parameter budgets.
The double-level variants use fewer
trainable parameters in each pair, with higher accuracy at the
low budget but lower accuracy at the medium and high budgets.
These results suggest that the benefits of deeper
reparameterization depend on the parameter budget.
Under tight budgets, the hierarchical parameterization may
enable more effective use of limited trainable parameters,
thereby improving accuracy.
At larger budgets, however, further factorization may impose
unnecessary structural constraints and make optimization more
difficult, limiting the model's ability to exploit the
available parameters and consequently reducing accuracy. }

\begin{table}[!t]
\caption{Ablation study of HiLoRA variants on LLaMA-3.1-8B commonsense reasoning. }
    \label{tab:hilora_variants_ablation}
    \centering
    \small
    \setlength{\tabcolsep}{4pt}
    \scalebox{0.90}{
    \begin{tabular}{lcccc}
    \toprule
    Variant & Params (M) & Avg. & W-Avg. & $\Delta$ W-Avg. \\
    \midrule
    HiLoRA A+B & 2.82 & 83.28 & 85.82 & 0.00 \\
    HiLoRA A-only & 2.83 & 82.22 & 84.79 & -1.03 \\
    HiLoRA B-only & 2.82 & 81.92 & 84.55 & -1.27 \\
    \bottomrule
    \end{tabular}}
        
\end{table}

\begin{table}[!t]
\caption{HiLoRA depth--budget ablation on LLaMA-3.1-8B (Commonsense-15K).}
\label{tab:hilora_multilevel_budget}
\centering
\footnotesize
\setlength{\tabcolsep}{2pt}
\renewcommand{\arraystretch}{1.08}
\begin{tabular*}{\columnwidth}{@{\extracolsep{\fill}}llccc@{}}
\toprule
Depth & Configuration & \shortstack{Params (M)} & Avg. & W-Avg. \\
\midrule
\multicolumn{5}{@{}l}{\emph{Low budget}} \\
Single & $r{=}16,\,k_1{=}8$ & 1.409 & 80.45 & 83.17 \\
Double & $r{=}16,\,k_1{=}16,\,k_2{=}16$ & 1.360 & 81.39 & 84.05 \\
\addlinespace[3pt]
\multicolumn{5}{@{}l}{\emph{Medium budget}} \\
Single & $r{=}16,\,k_1{=}12$ & 2.114 & 82.07 & 84.84 \\
Double & $r{=}16,\,k_1{=}16,\,k_2{=}24$ & 2.040 & 81.70 & 84.03 \\
\addlinespace[3pt]
\multicolumn{5}{@{}l}{\emph{High budget}} \\
Single & $r{=}16,\,k_1{=}16$ & 2.818 & 83.28 & 85.82 \\
Double & $r{=}16,\,k_1{=}16,\,k_2{=}32$ & 2.720 & 81.75 & 84.42 \\
\bottomrule
\end{tabular*}
\par\smallskip
\begin{minipage}{\columnwidth}\footnotesize\raggedright
\end{minipage}
\end{table}

\paragraph{Reshaping Strategy}


We study different reshaping strategies in HiLoRA. Specifically, we evaluate our current Near-Square strategy: a row/column transfer strategy, which moves complete rows or columns, depending on which dimension is longer, to the shorter dimension. We also evaluate the Random mean strategy, which randomly permutes all parameters and reshapes them into an approximately square matrix, with the results averaged over three independent runs. The Elongated 1:4 strategy follows a procedure similar to Near-Square reshaping, except that the parameters are reshaped into a matrix with an aspect ratio close to 1:4.
The Elongated 1:16 strategy follows the same procedure as the Elongated 1:4 strategy, but adopts a more elongated matrix shape with a larger aspect ratio. 

As shown in Table~\ref{tab:reshaping_ablation_summary}, the Near-Square reshaping strategy achieves better performance than the other reshaping strategies. This verifies that reshaping trainable parameters into approximately square matrices provides a more effective low-rank parameterization. Further details are provided in Appendix~\ref{sec:reshape-strategy}.

\begin{table}[!t]
\caption{Reshaping ablation on LLaMA-3.1-8B commonsense reasoning. }
     \label{tab:reshaping_ablation_summary}

    \centering
    \small
    \scalebox{0.88}{\begin{tabular}{lcccc}
        \toprule
        Reshaping strategy & Params(M) & Avg. & W-Avg. & $\Delta$ W-Avg. \\
        \midrule
        Near-Square & 2.82 & 83.28 & 85.82 & 0.00 \\
        Random mean & 2.82 & 80.95 & 83.65 & -2.17 \\
        \midrule
        Elongated 1:4 & 2.82 & 81.02 & 83.40 & -2.42 \\
        Elongated 1:16 & 2.93 & 81.04 & 83.55 & -2.27 \\
        \bottomrule
    \end{tabular}}

\end{table}

\paragraph{Outer Rank $r$ and Hierarchical Rank $k$}

We investigate the effect of the outer rank $r$ and hierarchical rank
$k$ in HiLoRA. The outer rank controls the adaptation capacity, while
the hierarchical rank determines the compression level. As shown in
Table~\ref{tab:hilora_rk_ablation_llama}, different combinations of $r$ and $k$
provide different trade-offs between parameter efficiency and
adaptation performance.



\begin{table}[!t]
\caption{Ablation study of the outer rank $r$ and hierarchical ranks $k_A,k_B$ for HiLoRA on LLaMA-3.1-8B commonsense reasoning. }
    \label{tab:hilora_rk_ablation_llama}
    \centering
    \small
    \setlength{\tabcolsep}{4pt}
    \scalebox{0.88}{\begin{tabular}{lcccc}
    \toprule
    Configuration & Params (M) & Avg. & W-Avg. & $\Delta$ W-Avg. \\
    \midrule
    $r{=}8,k_A{=}24,k_B{=}20$   & 2.86 & 82.01 & 84.30 & -1.52 \\
    $r{=}16,k_A{=}16,k_B{=}16$  & 2.82 & 83.28 & 85.82 & 0.00 \\
    $r{=}24,k_A{=}14,k_B{=}12$  & 2.83 & 81.79 & 84.39 & -1.43 \\
    $r{=}32,k_A{=}12,k_B{=}10$  & 2.86 & 81.94 & 84.33 & -1.49 \\
    $r{=}64,k_A{=}8,k_B{=}8$    & 2.82 & 82.52 & 85.04 & -0.78 \\
    \bottomrule
    \end{tabular}}   
\end{table}


\section{Conclusion and Discussion}
\label{sec:conclusion}

In this work, we introduced HiLoRR, a hierarchical low-rank reparameterization framework for further compressing the trainable parameters introduced by parameter-efficient fine-tuning methods.
HiLoRR reorganizes the parameters generated by an existing PEFT
method into approximately square matrices and represents them using
additional low-rank factorizations. This operation can be applied
recursively, forming a hierarchical parameterization.

Our main finding is that the trainable parameters introduced by existing PEFT methods contain substantial internal redundancy. When applied to LoRA, HiLoRA retains the original outer adaptation rank while using only about one tenth of the trainable parameters of the full-size LoRA adapter. 
Despite this substantial reduction, the compressed models maintain performance close to, and in some settings comparable with, the original methods across different model architectures and reasoning tasks. 
These results demonstrate that effective downstream adaptation can be achieved with a considerably more compact parameterization, making HiLoRR particularly suitable when adapter storage, optimizer states, or task-specific parameter budgets are constrained.

Moreover, applying the same reparameterization principle to LoRA, DoRA, MoSA, and RaSA suggests that our method provides a general compression framework rather than a method-specific modification.

The additional multiplicative structure changes the optimization behavior and makes performance more sensitive to the learning rate, factor initialization, hierarchical rank, and reshaping strategy. Future work may explore adaptive rank allocation, architecture-aware reshaping, and more efficient optimization strategies, as well as extending HiLoRR to broader model families, modalities, and downstream tasks.

\FloatBarrier
\bibliographystyle{IEEEtran}
\bibliography{src/reference}

\FloatBarrier
\clearpage
\appendix
\numberwithin{equation}{section}
\setcounter{table}{0}
\setcounter{figure}{0}
\renewcommand{\thetable}{S\arabic{table}}
\renewcommand{\thefigure}{S\arabic{figure}}


\section{Additional Details of HiLoRR}

\subsection{Reshaping Strategy}
\label{sec:reshape-strategy}

Fig.~\ref{fig:hilora_near_square_reshape}(a) illustrates
the near-square reshaping operation.
For a LoRA factor $X\in\mathbb R^{m\times n}$, let $N=mn$.
We select the largest divisor $p$ of $N$ not exceeding
$\lfloor\sqrt{N}\rfloor$ and set $q=N/p$, obtaining the exact
divisor pair closest to square.
The factor is flattened in row-major order and reshaped into
a $p\times q$ matrix via $\mathrm{reshape}(\cdot)$, preserving
the element order while changing the row boundaries.
For example, the illustrated $2\times8$ factor becomes a
$4\times4$ matrix.
This operation involves no transposition, permutation,
padding, or truncation; the numbers in the figure indicate
element positions rather than weight values.

Fig.~\ref{fig:hilora_near_square_reshape}(b) shows the
corresponding forward computation.
Rather than optimizing $X$ independently, we directly learn
$U\in\mathbb R^{p\times k}$ and
$V\in\mathbb R^{k\times q}$ and reconstruct the factor as
\[
    X=\text{Back}(UV),
\]
where $\text{Back}(\cdot)$ restores the original dimensions
using the same row-major element order.
Thus, reshaping itself is an invertible layout transformation;
parameter savings arise from learning the $k(p+q)$ entries
of $U$ and $V$ instead of the $mn$ entries of $X$,
provided that $k(p+q)<mn$.

\usetikzlibrary{arrows.meta}
\begin{figure*}[t]
\centering
\resizebox{0.98\textwidth}{!}{%
\begingroup
\definecolor{rsInk}{HTML}{263548}
\definecolor{rsGrid}{HTML}{91A0AE}
\definecolor{rsBlue}{HTML}{DBEAFE}
\definecolor{rsOrange}{HTML}{FCE2C6}
\definecolor{rsGreen}{HTML}{D7EDE2}
\definecolor{rsViolet}{HTML}{E9E0F4}
\begin{tikzpicture}[
    x=1cm,y=1cm,
    font=\sffamily\small,
    text=rsInk,
    line cap=round,line join=round,
    flow/.style={-{Latex[length=2mm]},draw=rsInk,line width=.8pt},
    cell/.style={draw=rsGrid,line width=.35pt},
    index/.style={font=\sffamily\fontsize{8}{9}\selectfont},
    heading/.style={font=\sffamily\bfseries\small,anchor=west},
    caption/.style={font=\sffamily\fontsize{9}{11}\selectfont,align=center},
    note/.style={font=\sffamily\fontsize{8}{10}\selectfont,align=center}
]

\node[heading] at (0,5.94)
    {(a) Near-square reshaping: preserve element order, change row boundaries};
\node[caption] at (1.84,5.23) {Original factor $X$\\$m\times n=2\times8$};
\node[caption] at (7.96,5.23) {Row-major sequence\\$N=mn=16$};
\node[caption] at (13.22,5.23) {Near-square layout $\mathrm{reshape}(X)$\\$p\times q=4\times4$};

\foreach \r in {0,1}{
    \ifnum\r=0\relax\def\rowfill{rsBlue}\else\def\rowfill{rsOrange}\fi
    \foreach \c in {0,...,7}{
        \pgfmathtruncatemacro{\idx}{8*\r+\c+1}
        \path[cell,fill=\rowfill]
            ({.46*\c},{4.26-.46*\r}) rectangle ++(.46,-.46);
        \node[index] at ({.46*\c+.23},{4.03-.46*\r}) {\idx};
    }
}
\draw[flow] (3.83,3.80) -- (4.85,3.80)
    node[midway,above=4pt,font=\sffamily\fontsize{8}{9}\selectfont] {flatten};

\foreach \c in {0,...,15}{
    \ifnum\c<8\relax\def\rowfill{rsBlue}\else\def\rowfill{rsOrange}\fi
    \pgfmathtruncatemacro{\idx}{\c+1}
    \path[cell,fill=\rowfill]
        ({5+.37*\c},4.03) rectangle ++(.37,-.46);
    \node[index] at ({5+.37*\c+.185},3.80) {\idx};
}
\draw[flow] (11.07,3.80) -- (12.15,3.80)
    node[midway,above=4pt,font=\sffamily\fontsize{8}{9}\selectfont] {reshape};

\foreach \r in {0,...,3}{
    \ifnum\r<2\relax\def\rowfill{rsBlue}\else\def\rowfill{rsOrange}\fi
    \foreach \c in {0,...,3}{
        \pgfmathtruncatemacro{\idx}{4*\r+\c+1}
        \path[cell,fill=\rowfill]
            ({12.30+.46*\c},{4.72-.46*\r}) rectangle ++(.46,-.46);
        \node[index] at ({12.53+.46*\c},{4.49-.46*\r}) {\idx};
    }
}

\node[note] at (7.15,2.29) {
    Shape selection:
    $p=\max\{d\in\mathbb N:\ d\mid N,\ d\leq\lfloor\sqrt N\rfloor\}$,
    $q=N/p$; here $p=q=4$.
};
\node[caption] at (7.15,1.71) {
    $[\mathrm{reshape}(X)]_{i,j}
    =X_{\lfloor(iq+j)/n\rfloor,\,(iq+j)\bmod n}$
    \quad (zero-based indices)
};
\node[note] at (7.15,1.17) {
    Numbers denote element positions. No transposition, permutation, padding, or truncation.
};

\draw[rsGrid!65,line width=.45pt] (0,.71) -- (14.60,.71);

\node[heading] at (0,.22)
    {(b) Actual forward computation: learn $U,V$, then reconstruct the factor};

\path[draw=rsGrid,fill=rsGreen,line width=.55pt]
    (.30,-.36) rectangle (1.20,-1.76);
\node at (.75,-1.06) {$U$};
\node[caption] at (.75,-2.24) {$U\in\mathbb R^{p\times k}$};
\node[font=\large] at (1.68,-1.06) {$\times$};
\path[draw=rsGrid,fill=rsViolet,line width=.55pt]
    (2.15,-.71) rectangle (4.15,-1.41);
\node at (3.15,-1.06) {$V$};
\node[caption] at (3.15,-2.24) {$V\in\mathbb R^{k\times q}$};
\node[note] at (2.23,-2.73) {Only $U$ and $V$ are directly optimized.};

\draw[flow] (4.40,-1.06) -- (5.51,-1.06)
    node[midway,above=4pt,font=\sffamily\fontsize{8}{9}\selectfont] {multiply};

\foreach \r in {0,...,3}{
    \ifnum\r<2\relax\def\rowfill{rsBlue}\else\def\rowfill{rsOrange}\fi
    \foreach \c in {0,...,3}{
        \path[cell,fill=\rowfill]
            ({5.75+.50*\c},{-.06-.50*\r}) rectangle ++(.50,-.50);
    }
}
\node[fill=white,inner sep=3pt] at (6.75,-1.06) {$UV$};
\node[caption] at (6.75,-2.24) {$Y=UV\in\mathbb R^{p\times q}$};

\draw[flow] (7.98,-1.06) -- (10.30,-1.06)
    node[midway,above=4pt,font=\sffamily\fontsize{8}{9}\selectfont] {reshape back};
\node[note] at (9.14,-1.67) {same row-major order};

\foreach \r in {0,1}{
    \ifnum\r=0\relax\def\rowfill{rsBlue}\else\def\rowfill{rsOrange}\fi
    \foreach \c in {0,...,7}{
        \path[cell,fill=\rowfill]
            ({10.45+.46*\c},{-.60-.46*\r}) rectangle ++(.46,-.46);
    }
}
\node[fill=white,inner sep=2pt] at (12.29,-1.06) {$X$};
\node[caption] at (12.29,-2.24)
    {$X=\text{Back}(UV)\in\mathbb R^{m\times n}$};

\node[note] at (7.15,-3.33) {
    \texttt{folded = U @ V}\quad$\longrightarrow$\quad
    \texttt{flat = folded.reshape(-1)}\quad$\longrightarrow$\quad
    \texttt{X = flat.reshape(m, n)}
};
\end{tikzpicture}
\endgroup
}
\caption{Near-square reshaping and factor reconstruction in HiLoRA.
(a) Element order is preserved while row boundaries change;
numbers indicate element positions rather than weight values.
(b) The factor is reconstructed as $X=\text{Back}(UV)$,
with only $U$ and $V$ directly optimized.}
\label{fig:hilora_near_square_reshape}
\end{figure*}

\subsection{HiLoRR Configurations}

HiLoRR can be configured with different reparameterization strategies
according to the target PEFT parameters. We consider three variants:
Single-side HiLoRR, Dual-side HiLoRR, and Multi-level HiLoRR.

\textbf{Single-side HiLoRR.}
Single-side HiLoRR applies hierarchical low-rank reparameterization to
only one trainable component while keeping the remaining parameters
unchanged. For LoRA-based methods, it only compresses one factor (e.g.,
$B$) and preserves the original update formulation.

\textbf{Dual-side HiLoRR.}
Dual-side HiLoRR extends Single-side HiLoRR by simultaneously
reparameterizing both trainable factors. For LoRA-based methods, both
$A$ and $B$ are compressed through HiLoRR, providing stronger parameter
reduction while maintaining the original adaptation mechanism.

\textbf{Multi-level HiLoRR.}
Multi-level HiLoRR recursively applies low-rank reparameterization to
the resulting factors. By increasing the hierarchy depth, HiLoRR
achieves additional parameter compression and provides a flexible
trade-off between compression ratio and adaptation capacity.

\section{Parameter Compression Analysis}\label{sec:paramanalysis}

\paragraph{Compression and Adaptation Capacity}
Because HiLoRR acts on the internal parameterization of trainable PEFT parameters, the reconstructed parameters retain the original interface of the underlying PEFT method. For the LoRA instantiation, the reconstructed $B$ and $A$ retain their shapes $m\times r$ and $r\times n$, so the update satisfies
\begin{equation}
\operatorname{rank}(\Delta W)\leq r.
\label{eq:rank-preserving}
\end{equation}
HiLoRR reduces the internal degrees of freedom of trainable PEFT parameters while maintaining the effective adaptation capability of the original PEFT formulation. 
This differs from directly reducing $r$, which lowers the maximum rank of the resulting update.

For a rank $k<\min(p,q)$, the factor matrices representable by Eq.~\eqref{eq:fold} form a structured subset of those allowed by the original PEFT formulation at the same outer rank $r$. 
When $k\geq\min(p,q)$, any $p\times q$ matrix can be represented as $UV$, and the parameterization recovers the full representational capacity of the original PEFT formulation factor. 

\paragraph{Compression ratio}
Consider a linear layer with $B\in\mathbb{R}^{m\times r}$ and $A\in\mathbb{R}^{r\times n}$. Standard LoRA optimizes both factors directly and therefore introduces
\begin{equation}
P_{\mathrm{LoRA}}=r(m+n)
\end{equation}
trainable parameters. 
Reparameterizing $B$ and $A$ as matrices of shapes $p_B\times q_B$ and $p_A\times q_A$ (with $p_Bq_B=mr$ and $p_Aq_A=rn$) and using hierarchical  ranks $k_B$ and $k_A$, the three structural variants require
\begin{align}
P_{B\text{-only}} &= k_B(p_B+q_B)+rn,\\
P_{A\text{-only}} &= mr+k_A(p_A+q_A),\\
P_{A+B} &= k_B(p_B+q_B)+k_A(p_A+q_A).
\label{eq:ff-parameters}
\end{align}
The double-sided variant compresses both factors, whereas each single-sided variant leaves one factor's $rn$ or $mr$ parameters intact, which then bounds the layer's overall compression.
Accordingly, we define the parameter retention ratio
\begin{equation}
\rho=\frac{{P_{A+B}}}{P_{\mathrm{LoRA}}},
\label{eq:retention-ratio}
\end{equation}
%
%
Per factor, the reshaped count is $k(p+q)$ instead of the $d$ entries of the original factor, corresponding to a compression factor $d/[k(p+q)]$. 
When the factor is reshaped into a nearly square matrix ($p\approx q\approx\sqrt{d}$), this is about $\frac{d}{k(p+q)} \approx \frac{\sqrt{d}}{2k}.$
%
%
The compression factor grows with the layer size and shrinks with the hierarchical rank $k$: larger layers yield greater compression, whereas larger hierarchical ranks yield less compression. 
%

\paragraph{Whole-model compression}
The per-factor compression factor above scales roughly with the layer size, 
so it varies across the heterogeneous layers of a Transformer, and the model-level saving is not simply the single-layer ratio. Summing over all adapted layers $\ell$, the original PEFT formulation and A+B 
HiLoRA use:
\begin{align}
P_{\mathrm{LoRA}}^{\mathrm{model}}&=\sum_{\ell} r(m_\ell+n_\ell),\\
P_{\mathrm{HiLoRA}}^{\mathrm{model}}&=\sum_{\ell}
\big[k_B(p_B^\ell+q_B^\ell)+k_A(p_A^\ell+q_A^\ell)\big]
\end{align}
trainable parameters, respectively, and the model-level parameter retention ratio is their quotient
\begin{equation}
\rho^{\mathrm{model}}
=\frac{P_{\mathrm{HiLoRA}}^{\mathrm{model}}}{P_{\mathrm{LoRA}}^{\mathrm{model}}}.
\label{eq:model-retention}
\end{equation}
Table~\ref{tab:reshape-shapes-llama-r16k16} instantiates this for LLaMA-3.1-8B, whose $32$ identical Transformer blocks share the same reshaping, 
so the per-block figures scale directly to the whole model.
Within one block, A+B HiLoRA compresses the adapted factors from $884{,}736$ to $88{,}064$ trainable parameters:
%
in the same-rank regime ($r16/k16$), HiLoRA uses about one tenth of the original PEFT parameters.

\section{Additional Experimental Details}

\subsection{Training Setup}

All experiments are conducted on a single NVIDIA H800 GPU. Models are trained for 3 epochs on the commonsense reasoning task and 2 epochs on the mathematical reasoning task. We use an effective batch size of 64 for both
the commonsense and mathematical reasoning experiments.




	

\subsection{Training Hyperparameters}

We provide the detailed training configurations used in our experiments.
Table~\ref{tab:hyperparams} summarizes the optimization settings for
different PEFT methods and tasks.

\begin{table}[t]
\caption{Common training hyperparameters. Target modules
		(\texttt{q,k,v,up,down\_proj}) are shared across all methods.}
	\label{tab:hyperparams}
    \centering
	\small
	\begin{tabular}{@{}lll@{}}
		\toprule
		Hyperparameter & Commonsense & Math \\
		\midrule
		Effective batch size    & 64 & 64 \\
		Micro-batch $\times$ accum. & $4\times16$ & $4\times16$ \\
		Epochs                  & 3    & 2    \\
		Max length              & 512  & 1024 \\
		Optimizer               & AdamW & AdamW \\
		HiLoRR $\alpha$        & 32   & 128  \\
		Dropout                 & 0.05 & 0.05 \\
		\bottomrule
	\end{tabular}
	
\end{table}

Unless otherwise specified, all methods use the same training
configuration to ensure a fair comparison. Task-dependent settings are summarized in Table~\ref{tab:hyperparams}.

\subsection{Parameter Compression Analysis}

We further provide detailed parameter compression analysis for HiLoRA.
Table~\ref{tab:reshape-shapes-llama-r16k16} reports the number of trainable
parameters before and after applying hierarchical low-rank
reparameterization.

\makeatletter
\setlength{\@fptop}{0pt}
\setlength{\@dblfptop}{0pt}
\makeatother
\begin{table*}[!t]
	\centering
	\caption{HiLoRA reshaping configurations on LLaMA-3.1-8B ($r{=}16$, $k{=}16$, recursive double-sided variant).
		Per Transformer block, summed over $\{$q,\,k,\,v,\,up,\,down$\}$\_proj.
		$d$: factor element count; std: the original PEFT formulation parameters; FF: reshaped parameters $k(p{+}q)$.}
	\label{tab:reshape-shapes-llama-r16k16}
	\begin{tabular}{llrcrrrrr}
		\toprule
		Module & Factor & Shape & $d$ & Reshape $(p{\times}q)$ & $k$ & std & FF & Ratio \\
		\midrule
		q\_proj    & $B$ (out) & $4096{\times}16$  & 65{,}536  & $256{\times}256$ & 16 & 65{,}536  & 8{,}192  & $8.0\times$ \\
		q\_proj    & $A$ (in)  & $16{\times}4096$  & 65{,}536  & $256{\times}256$ & 16 & 65{,}536  & 8{,}192  & $8.0\times$ \\
		k\_proj    & $B$ (out) & $1024{\times}16$  & 16{,}384  & $128{\times}128$ & 16 & 16{,}384  & 4{,}096  & $4.0\times$ \\
		k\_proj    & $A$ (in)  & $16{\times}4096$  & 65{,}536  & $256{\times}256$ & 16 & 65{,}536  & 8{,}192  & $8.0\times$ \\
		v\_proj    & $B$ (out) & $1024{\times}16$  & 16{,}384  & $128{\times}128$ & 16 & 16{,}384  & 4{,}096  & $4.0\times$ \\
		v\_proj    & $A$ (in)  & $16{\times}4096$  & 65{,}536  & $256{\times}256$ & 16 & 65{,}536  & 8{,}192  & $8.0\times$ \\
		up\_proj   & $B$ (out) & $14336{\times}16$ & 229{,}376 & $448{\times}512$ & 16 & 229{,}376 & 15{,}360 & $14.9\times$ \\
		up\_proj   & $A$ (in)  & $16{\times}4096$  & 65{,}536  & $256{\times}256$ & 16 & 65{,}536  & 8{,}192  & $8.0\times$ \\
		down\_proj & $B$ (out) & $4096{\times}16$  & 65{,}536  & $256{\times}256$ & 16 & 65{,}536  & 8{,}192  & $8.0\times$ \\
		down\_proj & $A$ (in)  & $16{\times}14336$ & 229{,}376 & $448{\times}512$ & 16 & 229{,}376 & 15{,}360 & $14.9\times$ \\
		\midrule
		\textbf{Total} & & & & & & \textbf{884{,}736} & \textbf{88{,}064} & $\mathbf{10.0\times}$ \\
		\bottomrule
	\end{tabular}
\end{table*}

The results show that HiLoRA substantially reduces the number of
trainable parameters while preserving the original adaptation structure.
Additional hierarchical reparameterization can yield further parameter savings
when the internal ranks are chosen to satisfy the compression condition.


\end{document}